\documentclass[sigconf,screen]{acmart}

\AtBeginDocument{%
  }

\usepackage{times}
\usepackage{soul}
\usepackage{url}
\usepackage{placeins}
\usepackage[utf8]{inputenc}
\usepackage{xspace}
\usepackage{graphicx}
\usepackage{svg}
\usepackage{amsmath}
\usepackage{amsthm}
\usepackage{booktabs}
\usepackage{algorithm}
\usepackage{algorithmic}
\usepackage[switch]{lineno}
\usepackage{xcolor}
\usepackage{natbib}
\definecolor{softpink}{HTML}{B85450}
\definecolor{softblue}{HTML}{6C8EBF}
\definecolor{softviolet}{HTML}{9673A6}

\usepackage{pdfpages}
\usepackage{booktabs}
\usepackage{multirow}
\usepackage{pifont}
\usepackage[dvipsnames]{xcolor}
\newcommand{\cmark}{\ding{51}}  
\newcommand{\xmark}{\ding{55}}  
\newcommand{\res}[1]{\ifx&#1& {---} \else #1 \fi}  

\acmConference[]{}{}{}

\setcopyright{none}
\renewcommand\footnotetextcopyrightpermission[1]{}

\begin{document}

\title{Dual-State Slot Attention: Decoupling Appearance and Identity for Video Object-Centric Learning}

\author{
Sieu Tran\textsuperscript{*},
Duc Nguyen\textsuperscript{*},
Hao Vo,
Khoa Vo,
Ngan Le
}

\thanks{\textsuperscript{*}Equal contribution.}

\affiliation{
\institution{University of Arkansas}
\city{Fayetteville}
\state{AR}
\country{USA}
}

\email{{stran5, dnguyen3, haov, khoavoho, thile}@uark.edu}

\renewcommand{\shortauthors}{Tran et al.}

\begin{abstract}
Unsupervised video object-centric learning aims to decompose dynamic scenes into persistent, object-level representations without supervision. However, existing slot-based methods struggle to maintain stable object identity in challenging settings such as rapid motion and partial occlusion. First, they typically encode both the per-frame appearance of an object and its identity across frames in a single slot vector, creating an objective conflict that leads to \emph{slot swapping}: reconstruction requires sensitivity to transient visual changes, whereas temporal consistency requires invariance to them. Second, the token renormalization used in Slot Attention can amplify weakly attending slots, allowing them to absorb tokens from other objects and destabilize slot-to-object correspondence.
We propose \textbf{Dual-State Slot Attention (DSSA)}, a fully self-supervised framework that addresses these limitations by separating appearance from identity and by reducing spurious updates from weakly matching slots. DSSA decomposes each slot into a \emph{local state} for per-frame appearance and an \emph{identity state} for temporally stable object information, thereby aligning reconstruction and temporal consistency with separate representations. The identity state is updated through a learned recurrent transition that acts as a temporal filter on the local state, while \emph{competition-modulated aggregation (CMA)} down-weights updates from weakly matching slots and prevents them from absorbing tokens from other objects. Experiments on MOVi-C, MOVi-D, and YouTube-VIS demonstrate that DSSA consistently improves segmentation quality and temporal consistency over prior methods, while also yielding stronger downstream object recognition and video dynamics prediction. Code is available at \href{https://uark-aicv.github.io/dssa/}{\texttt{uark-aicv.github.io/dssa}}.
\end{abstract}



\maketitle

\begin{figure}[!t]
\centering
\includegraphics[width=0.95\linewidth]{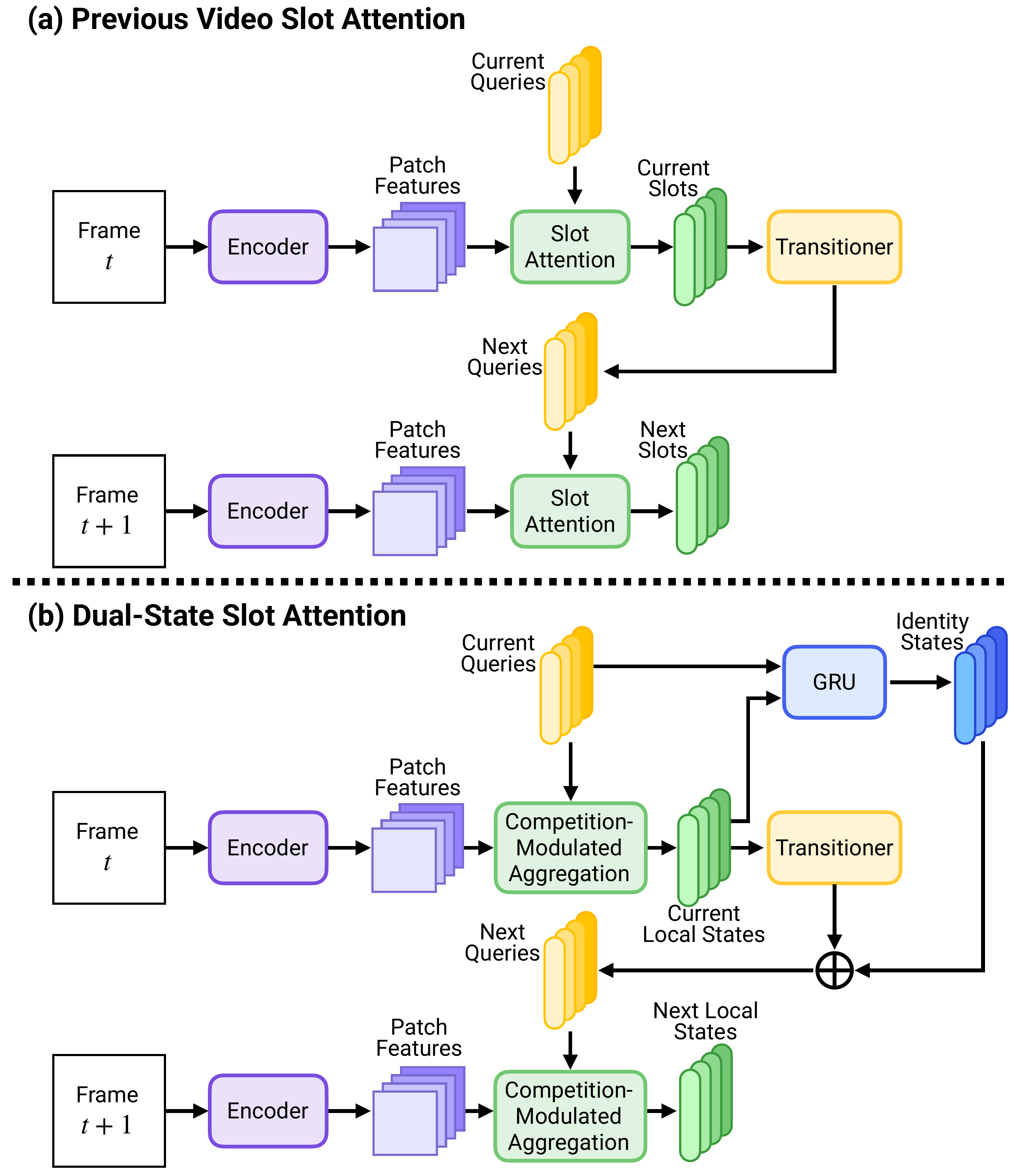}
\vspace{-3mm}
\caption{\textbf{Comparison of video object-centric learning (OCL) approaches.}
(a)~Prior video-based Slot Attention methods encode both appearance and identity 
within a single slot vector, causing reconstruction and temporal 
consistency to compete over a shared representation.
(b)~Our proposed DSSA assigns each slot a dedicated \emph{local state} for
per-frame reconstruction and an \emph{identity state} for temporally stable
object tracking, resolving this conflict by design.}
\label{fig:teaser}
\vspace{-6mm} 
\end{figure}

\section{Introduction}

Humans perceive the visual world as a structured collection of discrete objects, each with its own properties and persistent identity. Object-centric learning (OCL)~\citep{burgess2019monet, greff2019multi, locatello2020object,aydemir2023self,didolkar2025ctrl} aims to endow neural networks with a similar capacity: decomposing a visual scene into modular, object-level representations without object-level supervision. This objective is motivated by a long-standing view in cognitive science and artificial intelligence: structured, compositional representations of discrete entities, rather than holistic feature maps, are fundamental to robust reasoning, scene understanding, and generalization to novel environments~\cite{lake2017building, bengio2013representation, didolkar2025ctrl, mamaghan2024exploring,greff2020binding,didolkar2024zero,lin2020space,yi2019clevrer,wang2025hierarchical}. Slot-based models~\cite{locatello2020object, seitzer2022bridging} operationalize this idea by encoding a scene into a small set of latent vectors called \emph{slots}, where each slot competes with others to attend to different spatial regions and jointly reconstruct the scene. On static image benchmarks, slot attention~\cite{seitzer2022bridging, kakogeorgiou2024spot, wu2023slotdiffusion,biza2023invariant,zhao2025dias} has demonstrated convincing unsupervised object discovery and has emerged as a dominant approach to OCL.

Extending slot-based OCL from images to video is a natural next step, as temporal signals provide rich cues for reinforcing persistent object identities across frames\cite{aydemir2023self,elsayed2022savi++,jiang2019scalor,kipf2021conditional,kabra2021simone}. However, video introduces a challenge absent in the static image setting: a slot must capture not just \emph{what} an object looks like in the current frame, but also \emph{which} object it corresponds to over time. While object appearance may vary substantially due to motion, deformation, viewpoint, or illumination, object identity must remain stable despite such changes. Prior video OCL methods have addressed this challenge through temporal propagation mechanisms -- such as recurrence, cross-frame attention, or learned slot transitions~\cite{kipf2021conditional, elsayed2022savi++, singh2022simple, wu2022slotformer, zadaianchuk2023object, manasyan2025temporally} -- and have demonstrated meaningful progress in tracking objects across frames. Building on these advances, we examine two structural aspects of existing slot-based video OCL that may contribute to remaining difficulties in maintaining stable object identity over time.

The first aspect is the \textbf{representational conflict} arising from a shared representational choice: appearance and identity are encoded within a \emph{single} latent vector per slot, as illustrated in Figure~\ref{fig:teaser}(a). This creates an inherent tension: reconstruction encourages sensitivity to transient per-frame appearance cues, while temporal consistency encourages insensitivity to them. Optimizing both objectives within a single representation leads to a conflicting optimization landscape; in practice, slots have been observed to track volatile appearance changes rather than maintain stable object identity, a phenomenon commonly referred to as \emph{slot swapping}~\cite{elsayed2022savi++, zadaianchuk2023object}, which becomes more pronounced under rapid motion or partial occlusion. The second aspect is the \textbf{renormalization artifact} inherent in the slot attention mechanism itself. After computing competitive attention, the standard mechanism renormalizes weights across tokens. While effective for static images, this can have unintended effects in dynamic videos: when an object becomes occluded or moves abruptly, its corresponding slot attends weakly, yet renormalization inadvertently amplifies these weak signals, forcing the slot to capture tokens belonging to other entities and disrupting slot-to-object correspondence.

Motivated by these observations, we propose Dual-State Slot Attention (DSSA), a framework that addresses these structural limitations at both the representational and mechanistic levels, as illustrated in Figure~\ref{fig:teaser}(b).
First, to resolve the objective conflict, DSSA equips each slot with two dedicated representations: a \emph{local state} which interacts directly with frame tokens and captures frame-specific appearance for reconstruction, and an \emph{identity state} which is updated through a learned recurrent transition to accumulate temporally stable object information. This recurrent update acts as a temporal filter that distills persistent object characteristics while suppressing frame-specific appearance variation. An auxiliary identity reconstruction loss further reinforces identity stability. This explicit factorization aligns reconstruction and temporal consistency with dedicated representations, eliminating the structural compromise inherent in single-vector models. Second, to mitigate the renormalization artifact, DSSA introduces \emph{competition-modulated aggregation (CMA)}. Instead of treating all renormalized slot-token assignments equally, this mechanism scales each slot’s aggregated representation by its competitive attention strength. As a result, slots with low competitive confidence effectively remain ``silent'' when no matching object is present, rather than being artificially amplified through token renormalization.  At the same time, slots with strong competitive support retain the benefits of balanced aggregation while preserving balanced activation for strongly competing slots. 

We evaluate DSSA on standard video OCL benchmarks spanning both synthetic and real-world datasets. DSSA achieves state-of-the-art performance relative to prior methods, including SlotContrast~\cite{manasyan2025temporally} and RandSF.Q~\cite{zhao2025randsfq}, with consistent gains in segmentation quality (e.g., +3.7 points FG-ARI on MOVi-D and +9.7 points on YouTube-VIS) and tracking stability across MOVi-C, MOVi-D, and YouTube-VIS. DSSA also demonstrates superior performance on downstream object recognition and video dynamics prediction. Ablation studies show that the two components are complementary and that their combination yields the most balanced profile across all metrics, pointing to the efficacy of separating appearance and identity at the architectural root.

\section{Related Work}\label{sec:related_work}

\subsection{Image Object-Centric Learning (OCL)}
\label{sec:related_image}
The foundations of slot-based OCL were established by generative 
models such as MONet~\cite{burgess2019monet}, 
IODINE~\cite{greff2019multi}, and GENESIS~\cite{engelcke2020genesis}, 
that decompose scenes into per-object representations via iterative 
inference. Slot Attention~\cite{locatello2020object} significantly 
simplified and unified this paradigm with a single differentiable 
module: $K$ slot vectors compete to attend to image tokens via a 
softmax normalized over slots, and the resulting attended values are 
then renormalized over tokens before updating each slot. This 
double-normalization is deliberate—the first softmax induces 
competition between slots, while the second normalization over tokens 
encourages each slot to receive a balanced share of the input, 
preventing any single slot from monopolizing all tokens. Together, 
these two operations form the core inductive bias for unsupervised 
object discovery and have become the standard design in the field.
Early slot-based methods trained with pixel reconstruction objectives 
were largely limited to synthetic datasets. Later, 
\cite{seitzer2022bridging} overcame this by replacing pixel targets 
with features from a frozen self-supervised ViT, demonstrating that 
slot-based discovery can scale to real-world images like 
COCO~\cite{lin2014microsoft}. Recently, 
SPOT~\cite{kakogeorgiou2024spot} further refined this with a 
student-teacher scheme. These advances establish the feature 
reconstruction objective that has since become the dominant training 
signal in video OCL, and they characterize the standard slot attention 
mechanism whose behavior in dynamic settings we revisit in this paper.
When slots are propagated across video frames, new demands arise: a 
slot must not only explain the current frame, but also maintain a 
consistent identity over time. DSSA revisits the standard slot 
formulation under this temporal setting and finds that both the 
single-vector slot representation and the token-level renormalization 
of Slot Attention introduce structural instabilities in video -- 
motivating a dual-state slot representation and a modified aggregation 
mechanism, described in Section~\ref{sec:method}.

\begin{figure*}[t]
\centering
\includegraphics[width=0.85\linewidth]{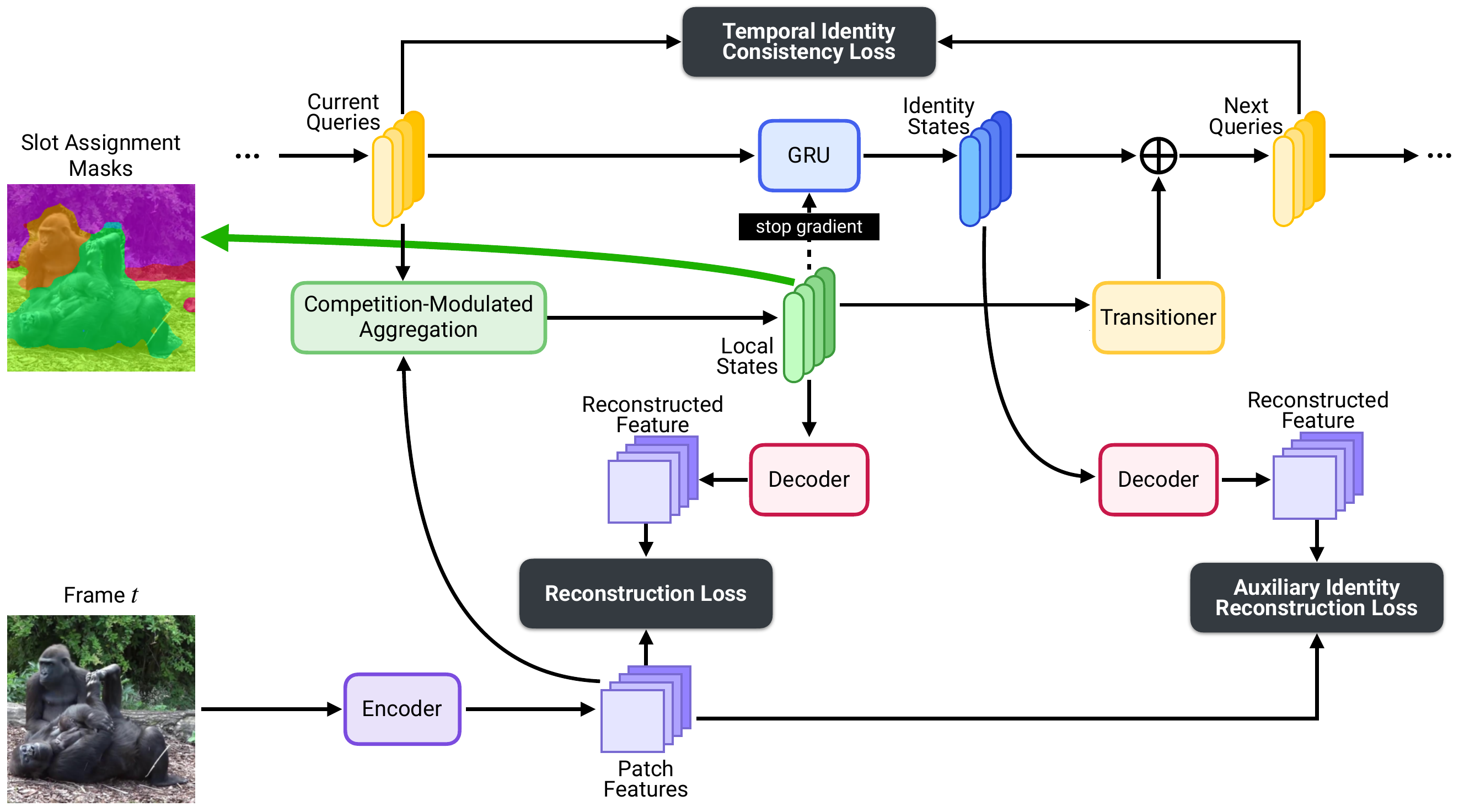}
\vspace{-3mm}
\caption{\textbf{Overview of Dual-State Slot Attention (DSSA).} At each frame, a frozen encoder extracts patch features, from which competition-modulated aggregation (CMA) produces local states, while slot assignment masks are derived from the slot-token attention assignments. The local states encode frame-specific appearance information, whereas identity states retain temporally stable object information distilled from the local states through a stop-gradient GRU. Combined with the transitioner output, these identity states form the next-frame queries. By separating local appearance from persistent identity, DSSA assigns frame-specific appearance and temporally stable object information to different latent states. Training uses reconstruction, auxiliary identity reconstruction, and temporal identity consistency losses.}
\label{fig:main}
\vspace{-4mm}
\end{figure*}

\subsection{Video Object-Centric Learning}
Extending slot-based OCL to video introduces a requirement absent from 
the image setting: each slot must maintain a consistent object identity 
across frames despite appearance changes. The dominant approach has 
been to propagate slot representations recurrently. 
SAVi~\cite{kipf2021conditional} used optical flow as a self-supervised 
target, while SAVi++~\cite{elsayed2022savi++} scaled this to 
real-world driving scenes. STEVE~\cite{singh2022simple} and 
SlotFormer~\cite{wu2022slotformer} leveraged transformer-based 
dynamics to predict future slot states, demonstrating that structured 
representations support complex downstream tasks. More recent work has 
improved temporal consistency through stronger training signals and 
better transition modules. VideoSAUR~\cite{zadaianchuk2023object} 
introduced a temporal similarity loss that encodes both semantic and 
motion information by training the model to predict patch-level 
feature similarities to a future frame, scaling slot-based discovery 
to diverse, unconstrained real-world videos. 
SlotContrast~\cite{manasyan2025temporally} introduced an object-level 
contrastive loss between slots of successive frames to explicitly 
enforce temporal consistency, yielding substantial improvements across 
synthetic and real-world benchmarks. RandSF.Q~\cite{zhao2025randsfq} 
further addressed limitations in the transition module by incorporating 
next-frame features into query prediction and training the transitioner 
on randomly sampled slot-feature pairs to better learn transition 
dynamics, achieving state-of-the-art performance on real-world video 
benchmarks.

Despite these advances, all of these methods retain the single-vector 
slot design, in which one representation must simultaneously support 
per-frame appearance reconstruction and temporally consistent object 
identity—two objectives that impose conflicting pressures on the same 
vector. The principle that stable identity and transient appearance 
should be represented separately is well-established in the broader 
literature: slow feature analysis~\cite{wiskott2002slow} formalizes 
that identity-level features vary more slowly than appearance-level 
features, a concept exploited by slow-fast 
architectures~\cite{feichtenhofer2019slowfast} in video understanding. 
In object tracking, separating an appearance descriptor from a 
persistent identity embedding is standard practice. Yet this separation 
has not been applied to unsupervised slot-based OCL, where a single 
vector per slot is universally used.

A further structural issue lies in the token-level renormalization of 
Slot Attention~\cite{locatello2020object}, which we show introduces 
instabilities when slots are propagated across frames. We analyze this 
mechanism in detail in Section~\ref{sec:background} and refer to its 
failure mode as the \textbf{renormalization artifact}. DSSA addresses 
both issues at the architectural level by decomposing each propagated 
slot into a \textbf{local state} for frame-specific appearance and an 
\textbf{identity state} for temporally persistent object information, 
and introducing \textbf{CMA} to resolve 
the renormalization artifact, as described in Section~\ref{sec:method}.

\section{Methodology}
\label{sec:method}

\subsection{Overview}
\label{sec:overview}

DSSA processes a video clip frame-by-frame by maintaining two dedicated representations for each of the $K$ slots: a \emph{local state} $\ell_t^k \in \mathbb{R}^{d}$, which encodes per-frame appearance, and an \emph{identity state} $e_t^k \in \mathbb{R}^{d}$, which accumulates a temporally stable object description, where $k \in \{1,\dots,K\}$
indexes the slot and $t$ indexes the timestep. Rather than encoding both objectives within a single vector, DSSA assigns each a dedicated gradient path: the local state is optimized exclusively by a reconstruction objective ($\mathcal{L}_{\mathrm{recon}}$), while the identity state is optimized by a contrastive consistency objective ($\mathcal{L}_{\mathrm{id}}$) and an auxiliary reconstruction signal ($\mathcal{L}_{\mathrm{aux}}$).

At each timestep $t$, the model operates in three stages. First, a frozen DINO encoder extracts patch tokens $X_t$ from frame $I_t$. Second, the spatial transitioner $\mathcal{T}$ adapts the previous local state to the current frame; the result is combined with the previous identity state to form the slot query, and CMA produces the updated local state $\ell_t^k$. Third, the identity state $e_t^k$ is updated from the recently updated local state $\ell_t^k$ via a GRU, with a stop-gradient on $\ell_t^k$ ensuring that no gradient from the identity objectives flows back into the local state. The full pipeline is illustrated in Figure~\ref{fig:main}.

\subsection{Slot Attention: Revisit and Motivation}
\label{sec:background}
Slot Attention~\cite{locatello2020object} maps a set of input token features $X = [x_1,\dots,x_N]^\top \in \mathbb{R}^{N \times d}$ to $K$ object-centric latent vectors, called \emph{slots}, through iterative cross-attention. Here, $x_n \in \mathbb{R}^{d}$ denotes the feature of the $n$-th token. Let $\{s^k\}_{k=1}^K$, with $s^k \in \mathbb{R}^{d}$, denote the current slot states. At each iteration, Slot Attention computes
a query vector $q^k = W_q s^k \in \mathbb{R}^{d}$ for slot $k$, and key and value vectors $k_n = W_k x_n \in \mathbb{R}^{d}$ and $v_n = W_v x_n \in \mathbb{R}^{d}$ for token $n$, where $W_q$, $W_k$, and $W_v$ are learned linear projections. The first normalization step
enforces competition across slots for each token:
\begin{equation}
    a_0^k[n] = \frac{
        \exp\!\left(\tfrac{1}{\sqrt{d}}\,(q^k)^\top k_n\right)
    }{
        \sum_{k'}\exp\!\left(\tfrac{1}{\sqrt{d}}\,(q^{k'})^\top k_n\right)
    }, \qquad \sum_k a_0^k[n] = 1 \;\forall\, n,
    \label{eq:sa_attn}
\end{equation}

where $a_0^k[n]$ is the raw competitive attention assigned by slot $k$ to token $n$, and $k'$ is a dummy index over slots. Thus, for each token, Eq.~\eqref{eq:sa_attn} produces a distribution over slots that reflects how strongly each slot claims that token. Slot Attention then performs a second normalization over tokens for each slot:
\begin{equation}
    a^k[n] = \frac{a_0^k[n]}{\sum_{n'} a_0^k[n'] + \epsilon},
    \label{eq:sa_renorm}
\end{equation}
where $n'$ is a dummy index over tokens and $\epsilon > 0$ is a small constant for numerical stability. Using these renormalized weights, the aggregated input to slot $k$ is computed as $ u^k = \sum_{n=1}^{N} a^k[n] \, v_n \;\in\; \mathbb{R}^{d}$, where $u^k$ is the token-aggregated update for slot $k$.

\begin{figure}[h]
\centering
\includegraphics[width=.8\linewidth]{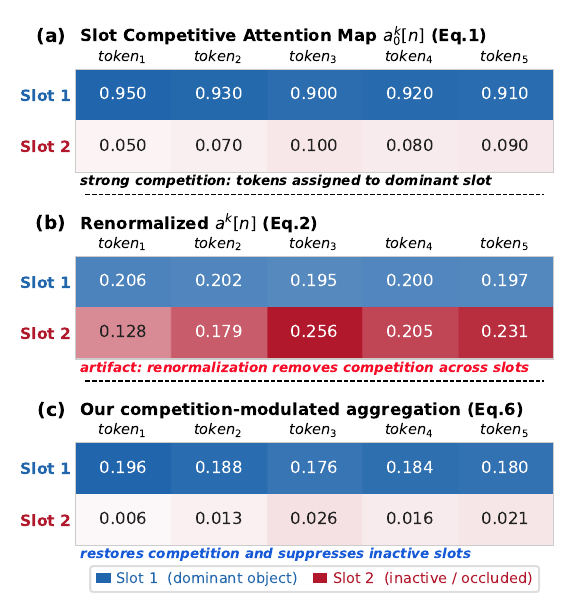}
\caption{\textbf{Analysis of Slot Attention Aggregation Weights.}
(a) Raw competitive attention $a_0^k[n]$: \textcolor{NavyBlue}{Slot 1} is dominant.
(b) Renormalized weights $a^k[n]$: \textcolor{BrickRed}{Slot 2} is amplified to match 
\textcolor{NavyBlue}{Slot 1}, erasing cross-slot competition.
(c) Our CMA preserves the original dominance structure.}
\label{fig:weight_analysis}
\vspace{-6mm} 
\end{figure}

This double normalization is designed to ensure balanced token coverage in static images. However, as illustrated in {Figure~\ref{fig:weight_analysis}}, it introduces a significant artifact in dynamic scenarios. In Figure~\ref{fig:weight_analysis} (a), \textcolor{NavyBlue}{Slot 1} correctly dominates the attention for all tokens, while \textcolor{BrickRed}{Slot 2} (representing an inactive or occluded object) receives very low raw scores (e.g., $0.050$). However, the renormalization in Eq.~\eqref{eq:sa_renorm} effectively removes this slot competitive information. As shown in Figure~\ref{fig:weight_analysis} (b), because the sum of weights for \textcolor{BrickRed}{Slot 2} is small, the division operation amplifies these weak signals, making \textcolor{BrickRed}{Slot 2} appear as active as \textcolor{NavyBlue}{Slot 1} ($0.128 \approx 0.194$).
In video sequences, when an object becomes temporarily occluded, its corresponding slot will receive uniformly weak competitive scores. Standard Slot Attention will then ``rescale'' these weak signals, causing the slot to absorb tokens that actually belong to other objects or the background. This effect accumulates over time, leading to a loss of slot-to-object consistency. We address this by introducing a CMA described in Section~\ref{sec:slot_attention}, which successfully suppresses inactive slots and preserves identity, as shown in Figure~\ref{fig:weight_analysis} (c).

\subsection{Feature Extraction}
\label{sec:feature}
 
At timestep $t$, a frozen Vision Transformer (ViT) pretrained with
DINO~\cite{caron2021emerging} maps frame $I_t$ to $N$ patch tokens:
\begin{equation}
    X_t = \mathrm{Enc}(I_t)\,W_p \;\in\; \mathbb{R}^{N \times d},
\end{equation}
where $W_p$ is a learned linear projection. The encoder is kept frozen throughout training. Following prior work~\cite{seitzer2022bridging, manasyan2025temporally}, we avoid optimizing the encoder jointly with a reconstruction objective. Doing so would shift the representations toward low-level pixel statistics, thereby disrupting the semantic grouping structure that slot attention relies on for object discovery. Instead, $W_p$ is used to adapt the frozen, semantics-rich features for object-centric grouping.

\subsection{Dual-State Slot Attention}
\label{sec:slot_attention}
 
\paragraph{Initialization.}
In DSSA, the slot state $s^k$ of standard Slot Attention is replaced by two dedicated states for each slot $k \in \{1,\dots,K\}$: a local state $\ell_t^k \in \mathbb{R}^{d}$ and an identity state $e_t^k \in \mathbb{R}^{d}$. At the first timestep, the identity states are initialized by sampling from a learned Gaussian distribution, following the standard Slot Attention initialization~\cite{locatello2020object}:
$e_0^k \sim \mathcal{N}(\mu, \mathrm{diag}(\sigma^2))$,
where $\mu \in \mathbb{R}^{d}$ and $\sigma \in \mathbb{R}^{d}$ are learned parameters shared across slots. The initial local state is set to zero: $\ell_0^k = \mathbf{0}$.

\paragraph{Query construction.}
At timestep $t$, the slot query $q_t^k$ combines two complementary signals. Unlike the standard query $W_q s^k$ in Eq.~\eqref{eq:sa_attn}, $q_t^k$ incorporates temporal information by fusing the previous identity state $e_{t-1}^k$, which carries a stable
description of \emph{which} object the slot is tracking, with a spatially adapted estimate of \emph{where} the object currently appears, produced by the spatial transitioner $\mathcal{T}$:
\begin{equation}
    q_t^k = e_{t-1}^k + \mathcal{T}(\ell_{t-1}^k,\, X_t)
    \;\in\; \mathbb{R}^{d},
    \label{eq:query}
\end{equation}
where $\mathcal{T}(\cdot,\cdot)$ denotes the spatial transitioner module, instantiated as a Transformer decoder following RandSF.Q~\cite{zhao2025randsfq}: $\ell_{t-1}^k$ serves as the query and $X_t$ serves as the key-value pairs. This design ensures that the slot
query is both temporally grounded by the identity state and spatially informed by the most recent local appearance, without requiring the identity state to attend directly to frame tokens at any point.

\paragraph{Competition-modulated aggregation (CMA).}
Given query $q_t^k$ and token features $X_t$, Slot Attention computes
raw competitive weights $a_{0,t}^k[n]$ (Eq.~\eqref{eq:sa_attn}) and
renormalized weights $a_t^k[n]$ (Eq.~\eqref{eq:sa_renorm}). To mitigate
the renormalization artifact described in Section~\ref{sec:background},
we modulate the renormalized weight $a_t^k[n]$ using the raw competitive
weight $a_{0,t}^k[n]$:
\begin{equation}
    \widetilde{a}_t^k[n]
    =
    a_t^k[n]\,
    \bigl(a_{0,t}^k[n]\bigr)^\alpha,
    \label{eq:modulated_aggregation}
\end{equation}

where $\alpha \geq 0$ controls the degree of competition modulation and interpolates 
between two limiting behaviors. When $\alpha = 0$, Eq.~\eqref{eq:modulated_aggregation} reduces to 
standard Slot Attention, preserving the renormalization artifact described in 
Section~\ref{sec:background}. When $\alpha = 1$, each slot's aggregation weight is scaled 
directly by its raw competitive score, which over-penalizes slots with moderate confidence 
and destabilizes updates. We set $\alpha = 0.5$, corresponding to a balanced interpolation 
between these two extremes: inactive slots are suppressed without imposing excessive 
penalties on moderately attending ones. This choice is further validated in 
Section~\ref{sec:ablation}, where $\alpha = 0.5$ achieves the best trade-off across all 
four metrics.
Note that
$\widetilde{a}_t^k[n]$ is \emph{not} renormalized again after
modulation. Instead, this formulation allows slots with low competitive
confidence -- such as one whose object is partially occluded -- to produce
weak updates and remain effectively silent when no reliable matching
object is present. The updated local state:
\begin{equation}
    \ell_t^k
    =
    \sum_{n=1}^{N} \widetilde{a}_t^k[n] \; W_v x_{t,n}
    \;\in\; \mathbb{R}^{d},
    \label{eq:local}
\end{equation}
where $x_{t,n} \in \mathbb{R}^{d}$ is the $n$-th token of $X_t$, and $W_v$ is the value projection defined in Section~\ref{sec:background}.
Consequently, the magnitude of $\ell_t^k$ naturally shrinks when the slot lacks confidence. This design ensures that the local state extracts frame-specific appearance information only from highly probable matches, effectively preventing unconfident slots from absorbing spurious evidence from other entities. The decoder implicitly learns to handle these dynamic variations in magnitude.

\subsection{Identity Update as Temporal Filtering}
\label{sec:identity}
 
The local state $\ell_t^k$ is optimized to capture frame-specific visual
evidence for reconstruction. While the local state may vary substantially
across time, the identity state $e_t^k$ is intended to encode only the
temporally persistent properties of the underlying object. To distill
such stable information from the local state, we update $e_t^k$ via a
gated recurrent unit (GRU) cell:

\begin{equation}
    e_t^k = \mathrm{GRUCell}\!\left(
        \mathrm{sg}(\ell_t^k),\; q_t^k
    \right),
    \label{eq:gru}
\end{equation}

where $\mathrm{sg}(\cdot)$ denotes stop-gradient, the detached local
state $\mathrm{sg}(\ell_t^k)$ serves as the GRU input, and the slot
query $q_t^k$ serves as the recurrent hidden state input to the GRU
cell. The GRU is applied for a single step per frame; $q_t^k$ thus
plays the role of the hidden state carried from the previous step,
warm-starting the update with both temporal context (via $e_{t-1}^k$)
and spatial information (via $\mathcal{T}(\ell_{t-1}^k, X_t)$), as
shown in Eq.~\eqref{eq:query}. The GRU then acts as a
temporal filter: it selectively retains slowly-varying identity features
from the new local state while discarding rapid appearance fluctuations. In this design, the identity state accumulates a temporally
stable representation of \emph{which} object the slot corresponds to,
whereas the local state remains specialized for \emph{what} the object
looks like in the current frame.
 
The stop-gradient on $\ell_t^k$ is the key mechanism that enforces this
separation. Specifically, the reconstruction loss
$\mathcal{L}_{\mathrm{recon}}$ optimizes $\ell_t^k$ through the
slot-attention pathway, shaping the local state to reconstruct the
current frame accurately. By contrast, the identity-related losses,
including $\mathcal{L}_{\mathrm{id}}$ and $\mathcal{L}_{\mathrm{aux}}$,
propagate through $e_t^k$ and the GRU parameters but are blocked at
$\mathrm{sg}(\ell_t^k)$. Thus, these losses shape how persistent
information is accumulated in the identity state without interfering with
the local state representation.

\subsection{Training Objectives}
\label{sec:objective}
 
The model is trained end-to-end without segmentation masks, object
tracks, or temporal correspondence labels, using three complementary
objectives that act on different components of the architecture.
 
\paragraph{Reconstruction loss.}
The local states $\{\ell_t^k\}_{k=1}^K$ are decoded by a shared
autoregressive Transformer decoder~\cite{zhao2025randsfq,zhao2025dias}
to reconstruct the current feature map $X_t$:
\begin{equation}
    \mathcal{L}_{\mathrm{recon}} = \frac{1}{NT}\sum_{t=1}^{T}\sum_{n=1}^{N}
        \bigl\|x_{t,n} - \hat{x}_{t,n}^{\ell}\bigr\|_2^2,
    \label{eq:recon}
\end{equation}
where $\hat{x}_{t,n}^{\ell} \in \mathbb{R}^{d}$ denotes the
reconstructed feature of token $n$ at timestep $t$ decoded from the
local states. This reconstruction objective acts exclusively on the local
state branch and shapes $\ell_t^k$ to capture per-frame appearance
details.
 
\paragraph{Auxiliary identity reconstruction loss.}
The same decoder independently reconstructs $\hat{x}_{t,n}^{e}$ from
the identity states $\{e_t^k\}_{k=1}^K$:
\begin{equation}
    \mathcal{L}_{\mathrm{aux}} = \frac{1}{NT}\sum_{t=1}^{T}\sum_{n=1}^{N}
        \bigl\|x_{t,n} - \hat{x}_{t,n}^{e}\bigr\|_2^2.
    \label{eq:aux}
\end{equation}
This auxiliary signal ensures that the identity state remains grounded
in the object's visual properties, preventing it from collapsing to a
degenerate solution under the contrastive loss alone.
 
\paragraph{Temporal identity consistency loss.}
We apply the slot-level contrastive loss~\cite{manasyan2025temporally} exclusively to the identity states. Given identity states at consecutive timesteps, we form a cosine similarity matrix across all slots and all samples in the batch, and supervise its softmax-normalized form toward the identity matrix via cross-entropy:
\begin{equation}
    \mathcal{L}_{\mathrm{id}} =
        \mathcal{L}_{\mathrm{CE}}\!\left(
            \mathrm{softmax}\!\left(
                \left\{
                    \frac{(e_t^k)^\top e_{t+1}^{k'}}
                         {\tau\,\|e_t^k\|\,\|e_{t+1}^{k'}\|}
                \right\}_{k,k'}
            \right),\; \mathbf{I}
        \right),
    \label{eq:id_loss}
\end{equation}
where $k$ and $k'$ each range over the $K$ slots across all $B$ samples
in the batch (forming a $KB \times KB$ similarity matrix), $\tau$ is a
temperature hyperparameter, $\mathbf{I}$ is the identity matrix, and
softmax is applied row-wise so that each slot at timestep $t$ is
matched to its corresponding slot at timestep $t{+}1$. Unlike SlotContrast, where this loss acts on
the full slot vector and therefore conflicts with the reconstruction
objective, here it acts exclusively on $e_t^k$, which is decoupled from
the local state by the stop-gradient in Eq.~\eqref{eq:gru}. This
eliminates the conflict by construction: $\mathcal{L}_{\mathrm{id}}$
can push the identity state toward pure temporal persistence without
compromising the per-frame reconstruction accuracy of the local state.

\paragraph{Full objective.}
The total loss combines all three objectives:
\vspace{-2mm}
\begin{equation}
    \mathcal{L} =
        \mathcal{L}_{\mathrm{recon}}
        + \frac{\mathcal{L}_{\mathrm{id}}
        + \mathcal{L}_{\mathrm{aux}}}{2},
    \label{eq:full}
\end{equation}
\begin{table*}[ht]
  \centering
  \caption{
    \textbf{Performance comparison with state-of-the-art video object-centric learning methods.}
    All methods use a frozen DINOv2 ViT-S/14 encoder and
    $256{\times}256$ input resolution.
    Results are mean~$\pm$~std over 3 seeds.
    \textbf{Bold}: best. \underline{Underlined}: second best.
  }
  \label{tab:main}
  \vspace{-3mm}
  \setlength{\tabcolsep}{5pt}
  \resizebox{\linewidth}{!}{%
  \begin{tabular}{l|cccc | cccc | cccc}
    \toprule
    \multirow{2}{*}{\textbf{Method}} & \multicolumn{4}{c|}{\textbf{MOVi-C} \small(\#slot=11, conditional)}
    & \multicolumn{4}{c|}{\textbf{MOVi-D} \small(\#slot=21, conditional)}
    & \multicolumn{4}{c}{\textbf{YTVIS} \small(\#slot=7)} \\
    \cmidrule(lr){2-5}\cmidrule(lr){6-9}\cmidrule(lr){10-13}
    
      & ARI & ARIfg & mBO & mIoU
      & ARI & ARIfg & mBO & mIoU
      & ARI & ARIfg & mBO & mIoU \\
    \midrule

    VideoSAUR~\cite{zadaianchuk2023object}
      & $41.9_{\pm1.1}$ & $53.3_{\pm2.1}$ & $16.1_{\pm0.4}$ & $14.8_{\pm0.4}$
      & $22.5_{\pm5.0}$ & $40.0_{\pm20.1}$ & $11.6_{\pm6.6}$ & $10.8_{\pm6.1}$
      & $33.8_{\pm0.7}$ & $49.2_{\pm0.5}$ & $29.9_{\pm0.4}$ & $29.7_{\pm0.4}$ \\
    SlotContrast~\cite{manasyan2025temporally}
      & $64.6_{\pm9.4}$ & $59.9_{\pm5.3}$ & $27.7_{\pm3.0}$ & $25.8_{\pm2.9}$
      & \underline{$45.3_{\pm4.1}$} & $63.9_{\pm0.2}$ & $26.7_{\pm1.0}$ & $25.1_{\pm1.0}$
      & $37.2_{\pm0.6}$ & $49.4_{\pm1.1}$ & $33.0_{\pm0.2}$ & $32.8_{\pm0.1}$ \\
    DIAS\textsubscript{video}~\cite{zhao2025dias}
      & $-$ & $-$ & $-$ & $-$
      & $37.2_{\pm3.5}$ & $64.7_{\pm3.7}$ & $25.9_{\pm2.4}$ & $22.7_{\pm2.6}$
      & $38.7_{\pm1.0}$ & $52.1_{\pm0.4}$ & $33.3_{\pm0.7}$ & $34.6_{\pm0.6}$ \\
    RandSF.Q\textsubscript{tsim}~\cite{zhao2025randsfq}
      & $64.0_{\pm2.9}$ & $66.3_{\pm1.7}$ & $28.4_{\pm1.3}$ & $26.1_{\pm1.1}$
      & $41.2_{\pm2.2}$ & $72.0_{\pm1.1}$ & $27.1_{\pm0.9}$ & $25.4_{\pm0.9}$
      & \underline{$46.0_{\pm0.7}$} & $60.4_{\pm2.3}$ & \underline{$39.4_{\pm0.3}$} & \underline{$38.5_{\pm0.2}$} \\
    RandSF.Q\textsubscript{ssc}~\cite{zhao2025randsfq}
      & \underline{$65.4_{\pm10.7}$} & $67.4_{\pm2.1}$ & $29.2_{\pm3.8}$ & $26.8_{\pm3.7}$
      & $41.6_{\pm3.7}$ & \underline{$77.5_{\pm1.0}$} & $27.4_{\pm1.0}$ & $25.6_{\pm1.0}$
      & $40.1_{\pm0.4}$ & $58.0_{\pm1.0}$ & $37.6_{\pm0.4}$ & $37.2_{\pm0.4}$ \\
    SmoothSA~\cite{zhao2026smoothsa}
      & $50.9_{\pm1.6}$ & $\mathbf{69.0_{\pm0.3}}$ & $\mathbf{31.7_{\pm0.8}}$ & $\mathbf{30.2_{\pm0.8}}$
      & $43.8_{\pm1.5}$ & $70.5_{\pm0.7}$ & $\mathbf{31.4_{\pm0.4}}$ & $\mathbf{30.2_{\pm0.4}}$
      & $42.4_{\pm0.8}$ & \underline{$63.0_{\pm3.4}$} & $38.9_{\pm0.7}$ & $38.3_{\pm0.6}$ \\
    \midrule
    \textbf{DSSA (ours)}
      & $\mathbf{67.6_{\pm9.1}}$ & \underline{$67.6_{\pm2.7}$} & \underline{$29.9_{\pm3.7}$} & \underline{$27.5_{\pm3.3}$}
      & $\mathbf{48.7_{\pm4.5}}$ & $\mathbf{81.2_{\pm1.8}}$ & \underline{$29.3_{\pm1.4}$} & \underline{$27.8_{\pm1.2}$}
      & $\mathbf{55.0_{\pm2.8}}$ & $\mathbf{70.1_{\pm1.5}}$ & $\mathbf{45.8_{\pm1.2}}$ & $\mathbf{44.7_{\pm1.1}}$ \\
    \bottomrule
  \end{tabular}}
\end{table*}
\noindent
where the identity-related losses are averaged equally. Each objective 
acts on a distinct component of the architecture -- 
$\mathcal{L}_{\mathrm{recon}}$ on the local state,
$\mathcal{L}_{\mathrm{id}}$ and $\mathcal{L}_{\mathrm{aux}}$ on the
identity state -- with the stop-gradient in Eq.~\eqref{eq:gru} ensuring
that these gradient paths remain structurally decoupled throughout
training.

\section{Experiment}
\label{sec:experiments}

\subsection{Experimental Setup}
\label{sec:setup}

\noindent
\textbf{Datasets.}
Following the experimental setup of RandSF.Q~\cite{zhao2025randsfq}, we evaluate our method on both synthetic and real-world video datasets. 
For synthetic datasets, we use \textit{MOVi-C} and \textit{MOVi-D}~\cite{greff2022kubric}, which feature everyday objects with complex textures on complex backgrounds, with MOVi-D being more challenging due to its larger number of objects per scene. 
For real-world evaluation, we use the high-quality (HQ) version of \textit{YouTube-VIS} (YTVIS)~\cite{Yang2019vis}, which contains diverse and complex~videos. 

\noindent
\textbf{Baselines.}
We compare against \textit{VideoSAUR}~\cite{zadaianchuk2023object},
\textit{SlotContrast}~\cite{manasyan2025temporally},
\textit{DIAS}\textsubscript{video}~\cite{zhao2025dias},
\textit{RandSF.Q}~\cite{zhao2025randsfq}, and
\textit{SmoothSA}~\cite{zhao2026smoothsa}.
Following RandSF.Q, we exclude SAVi~\cite{kipf2021conditional} and
SAVi++~\cite{elsayed2022savi++}, which require external supervision.

\noindent
\textbf{Metrics.} We report four video-level metrics computed over full video sequences,
which jointly reflect object discovery quality and temporal
consistency.
\textit{ARI} (Adjusted Rand Index) and \textit{ARIfg} (foreground
ARI)~\cite{greff2019multi} measure how consistently objects are
segmented across the full video; computing them at the video level
penalizes slot swaps and identity drift between frames.
\textit{mBO} (mean Best Overlap)~\cite{seitzer2022bridging} measures
mask sharpness via best-matched overlap between predicted and
ground-truth segments.
\textit{mIoU} (mean Intersection over Union~\cite{seitzer2022bridging})
provides a stricter spatial accuracy measure.

\noindent
\textbf{Implementation details.}
 For fair comparison with prior work, we follow the experimental protocol of RandSF.Q~\cite{zhao2025randsfq} where appropriate. We use a frozen DINOv2 ViT-S/14~\cite{oquab2023dinov2} encoder with 
a learned projection $W_p$, a single Transformer decoder block as 
spatial transitioner $\mathcal{T}$, and an autoregressive Transformer 
decoder for reconstruction. All models use $256{\times}256$ input and 
slots $K \in \{11, 21, 7\}$ for MOVi-C, MOVi-D, and YTVIS. 
We set $\alpha{=}0.5$ for CMA (validated in 
Section~\ref{sec:ablation}).
\subsection{Main Results}
\label{sec:discovery}

Table~\ref{tab:main} compares DSSA with prior video object-centric learning methods across synthetic and real-world benchmarks. DSSA achieves the best ARI on all three benchmarks and the best or second-best result on every metric. On MOVi-C, DSSA improves over RandSF.Q\textsubscript{ssc} by $+2.2$ ARI, $+0.2$ ARIfg, $+0.7$ mBO, and $+0.7$ mIoU, while the concurrent SmoothSA~\cite{zhao2026smoothsa} attains higher ARIfg, mBO, and mIoU but a substantially lower ARI ($50.9$ vs.\ $67.6$), indicating weaker temporal consistency of slot assignments. On the more challenging MOVi-D benchmark, DSSA achieves the best ARI and ARIfg ($+3.4$ ARI and $+3.7$ ARIfg over the best baseline), with SmoothSA leading on mBO and mIoU; here scenes contain more objects and stronger visual ambiguity -- a setting where consistent slot-to-object correspondence across frames becomes increasingly important for accurate segmentation. The improvements are largest on the real-world YTVIS benchmark, where DSSA is best on all four metrics ($+9.0$ ARI, $+7.1$ ARIfg, $+6.4$ mBO, $+6.2$ mIoU over the best baseline), suggesting that the dual-state design and CMA are particularly beneficial under severe appearance variation, background clutter, and occlusion.
Notably, the gains in ARI -- which, computed at the video level, directly penalizes slot swaps -- across all three benchmarks reflect the advantage of separating local appearance from persistent slot identity, as slots can maintain stable object associations even when visual features change significantly between frames.
Overall, the performance improvements become more pronounced as scene complexity increases, indicating that DSSA benefits both spatial grouping quality and the stability of slot assignments over time.
\vspace{-2mm}
\subsection{Downstream Tasks}
\label{sec:downstream}

We further evaluate the quality of the learned representations on two downstream tasks on YTVIS, assessing the local state $\ell_t^k$ and identity state $e_t^k$ separately to verify that each has specialized as intended.

\noindent
\textbf{Object recognition.}
Following RandSF.Q~\cite{zhao2025randsfq}, we freeze the OCL model
and train a two-layer MLP to predict object class and bounding box
from each slot representation using YTVIS annotations.
We compare against SlotContrast and RandSF.Q.
Since object classification and localization depend on fine-grained per-frame spatial detail, this task primarily evaluates the quality of the local visual information, which is explicitly optimized by $\mathcal{L}_{\mathrm{recon}}$. As shown in Table~\ref{tab:recognition}, the local state $\ell_t^k$ achieves the best object recognition performance, reaching \textbf{28.4} Top-1 and \textbf{66.6} Top-3 accuracy, outperforming both RandSF.Q and SlotContrast. This result is consistent with the design of DSSA: the local state is directly optimized by $\mathcal{L}_{\mathrm{recon}}$ to preserve frame-specific appearance cues that are most relevant for recognition. Bbox IoU remains comparable to RandSF.Q (52.1 vs.\ 54.5), while the number of matched samples is substantially higher (9269 vs.\ 7579), indicating broader object coverage across frames.
Note that after our experiments were conducted, RandSF.Q~\cite{zhao2025randsfq} released an updated evaluation protocol for this downstream task. Table~\ref{tab:recognition} follows the original protocol; results under the updated protocol are reported in the supplementary material (Table~S-VIII).

\begin{table}[t]
  \centering
  \caption{
    \textbf{Object recognition on YTVIS.}
    Two-layer MLP trained on frozen representations, following the
    original RandSF.Q protocol. Results under the updated RandSF.Q
    protocol are reported in the supplementary (Table~S-VIII).
    Best in \textbf{bold}.
  }
  \label{tab:recognition}
  \vspace{-3mm}
  \resizebox{.45\textwidth}{!}{%
  \begin{tabular}{l|cccc}
    \toprule
    Method & Top-1$\uparrow$ & Top-3$\uparrow$ & bbox IoU$\uparrow$ & match$\uparrow$ \\
    \midrule
    SlotContrast~\cite{manasyan2025temporally} & 19.9{\scriptsize$\pm$2.0} & 49.1{\scriptsize$\pm$3.1} & 53.5{\scriptsize$\pm$0.2} & 9259{\scriptsize$\pm$26} \\
    RandSF.Q\textsubscript{tsim}~\cite{zhao2025randsfq}            & 26.1{\scriptsize$\pm$1.3} & 60.9{\scriptsize$\pm$3.2} & \textbf{54.5{\scriptsize$\pm$0.6}} & 7579{\scriptsize$\pm$201} \\
    \midrule
    DSSA (identity $e_t^k$)          & 24.2{\scriptsize$\pm$0.9} & 63.2{\scriptsize$\pm$0.8} & 50.7{\scriptsize$\pm$0.3} & \textbf{9292{\scriptsize$\pm$26}} \\
    \textbf{DSSA (local $\ell_t^k$)} & \textbf{28.4{\scriptsize$\pm$2.9}} & \textbf{66.6{\scriptsize$\pm$2.0}} & 52.1{\scriptsize$\pm$1.1} & 9269{\scriptsize$\pm$14} \\
    \bottomrule
  \end{tabular}}
\end{table}

\noindent
\textbf{Object dynamics prediction.}
Following SlotContrast~\cite{manasyan2025temporally}, we train
SlotFormer~\cite{wu2022slotformer} on top of frozen slot
representations for object dynamics prediction.
SlotFormer predicts slots autoregressively for $K$ rollout steps based
on slots inferred from $T$ burn-in frames preceding the prediction
horizon.
Both the OCL model and SlotFormer operate in feature
space; we use only the slot reconstruction loss when training
SlotFormer.
We compare against SlotContrast and RandSF.Q.
Since accurate dynamics modeling requires representations that are stable
and consistent across frames, this task primarily evaluates the quality of the identity state, which is trained by $\mathcal{L}_{\mathrm{id}}$.

Table~\ref{tab:dynamics} shows that the identity state $e_t^k$ achieves the best performance across all metrics, obtaining \textbf{66.6} ARI, \textbf{52.3} ARIfg, \textbf{52.7} mBO, and \textbf{51.4} mIoU. These gains confirm that the identity state provides a more temporally stable representation for forecasting future object states, which aligns with its training objective $\mathcal{L}_{\mathrm{id}}$.

\begin{table}[t]
  \centering
  \caption{
    \textbf{Object dynamics prediction on YTVIS.}
    SlotFormer~\cite{wu2022slotformer} trained on frozen
    representations. Best in \textbf{bold}.
  }
  \label{tab:dynamics}
  \vspace{-3mm}
  \resizebox{.45\textwidth}{!}{%
  \begin{tabular}{l|cccc}
    \toprule
    Method
      & ARI$\uparrow$ & ARIfg$\uparrow$ & mBO$\uparrow$ & mIoU$\uparrow$ \\
    \midrule
    SlotContrast~\cite{manasyan2025temporally} & 37.9{\scriptsize$\pm$0.1} & 29.5{\scriptsize$\pm$0.2} & 33.2{\scriptsize$\pm$0.1} & 33.1{\scriptsize$\pm$0.1} \\
    RandSF.Q\textsubscript{tsim}~\cite{zhao2025randsfq}            & 46.6{\scriptsize$\pm$0.1} & 38.2{\scriptsize$\pm$0.5} & 43.7{\scriptsize$\pm$0.1} & 43.1{\scriptsize$\pm$0.2} \\
    \midrule
    DSSA (local $\ell_t^k$)                   & 66.5{\scriptsize$\pm$0.1} & 51.5{\scriptsize$\pm$0.2} & 52.6{\scriptsize$\pm$0.0} & 51.2{\scriptsize$\pm$0.1} \\
    \textbf{DSSA (identity $e_t^k$)}          & \textbf{66.6{\scriptsize$\pm$0.1}} & \textbf{52.3{\scriptsize$\pm$0.2}} & \textbf{52.7{\scriptsize$\pm$0.1}} & \textbf{51.4{\scriptsize$\pm$0.1}} \\
    
    \bottomrule
  \end{tabular}}
  \vspace{-4pt} 
\end{table}

\vspace{-2mm}
\subsection{Ablation Studies}
\label{sec:ablation}

\begin{table}[t]
  \centering
  \caption{
    \textbf{Component ablation on MOVi-C}.
    ``Dual'' = dual-state local/identity decomposition;
    ``CMA'' = competition-modulated aggregation ($\alpha=0.5$).
    Best in \textbf{bold}, second best \underline{underlined}.
  }
  \label{tab:ablation}
  \vspace{-3mm}
  \setlength{\tabcolsep}{3pt}
  \begin{tabular}{c|cc|cccc|cc}
    \toprule
    Exp.&Dual & CMA
      & ARI & ARIfg & mBO & mIoU & Params (M) & FPS \\
    \midrule
   \#1 & \xmark & \xmark
     &  53.3 & 65.5 & 24.7 & 22.5 & 34.1 & ~400 \\
   \#2 & \xmark & \cmark
     & \textbf{71.2} & 57.6 & 26.7 & 23.7 & 34.1 & ~400 \\
  \#3  &  \cmark & \xmark
    & 57.7 & \textbf{70.2} & 26.1 & 24.1 & 34.5 & ~380 \\
  \#4  &  \cmark & \cmark
    & \underline{68.6} & \underline{67.6} & \textbf{28.9} & \textbf{26.5} 
      & 34.5 & ~380 \\
    \bottomrule
  \end{tabular}
  \vspace{-8pt} 
\end{table}

\noindent
\textbf{Component contribution.} Table~\ref{tab:ablation} evaluates the 
effect of the two main design components: the dual-state decomposition 
and CMA ($\alpha = 0.5$). Both components contribute positively and are complementary.
CMA alone (Exp.\#1 v.s. \#2) substantially improves ARI ($53.3 \to 71.2$) but causes 
ARIfg to drop ($65.5 \to 57.6$), indicating that without a dedicated 
identity representation, slots achieve cleaner spatial grouping but 
struggle to maintain consistent object assignments.
The dual-state design alone (Exp.\#1 vs \#3) improves ARIfg ($65.5 \to 70.2$) with 
moderate gains elsewhere, confirming that separating appearance from 
identity provides a stronger inductive bias for stable slot-to-object 
correspondence.
Their combination (Exp.\#4) resolves the ARIfg degradation seen with CMA alone 
and achieves the most balanced profile, with minimal 
computational overhead: CMA adds no parameters, and the dual-state 
decomposition adds only $0.4$M parameters ($34.1 \to 34.5$M) with 
a modest throughput reduction (${\sim}400 \to {\sim}380$ FPS).
Note that the full model is not the best on every individual metric: CMA
alone yields the highest ARI and the dual-state design alone the highest
ARIfg. However, these peaks are one-sided: CMA alone loses $7.9$ ARIfg
relative to the baseline (Exp.\#1), and the dual-state design alone gains
only $4.4$ ARI. In contrast, Exp.\#4 is the only configuration that ranks
best or second best on every metric, and it achieves the best mBO and
mIoU. Since each
metric has its own blind spot that degenerate predictions can exploit
(see supplementary Sec.~2, Table~S-II), we regard this balanced profile,
rather than a peak on any single metric, as the more reliable indicator
of segmentation quality.
Further ablations of the training objectives ($\mathcal{L}_{\mathrm{id}}$,
$\mathcal{L}_{\mathrm{aux}}$) and of the stop-gradient in
Eq.~\eqref{eq:gru} are reported in the supplementary material
(Tables~S-III and~S-IV), and a targeted evaluation on sequences with
full occlusion and re-emergence is given in Table~S-V.

\begin{figure*}[h]
  \centering
  \includegraphics[width=\linewidth]{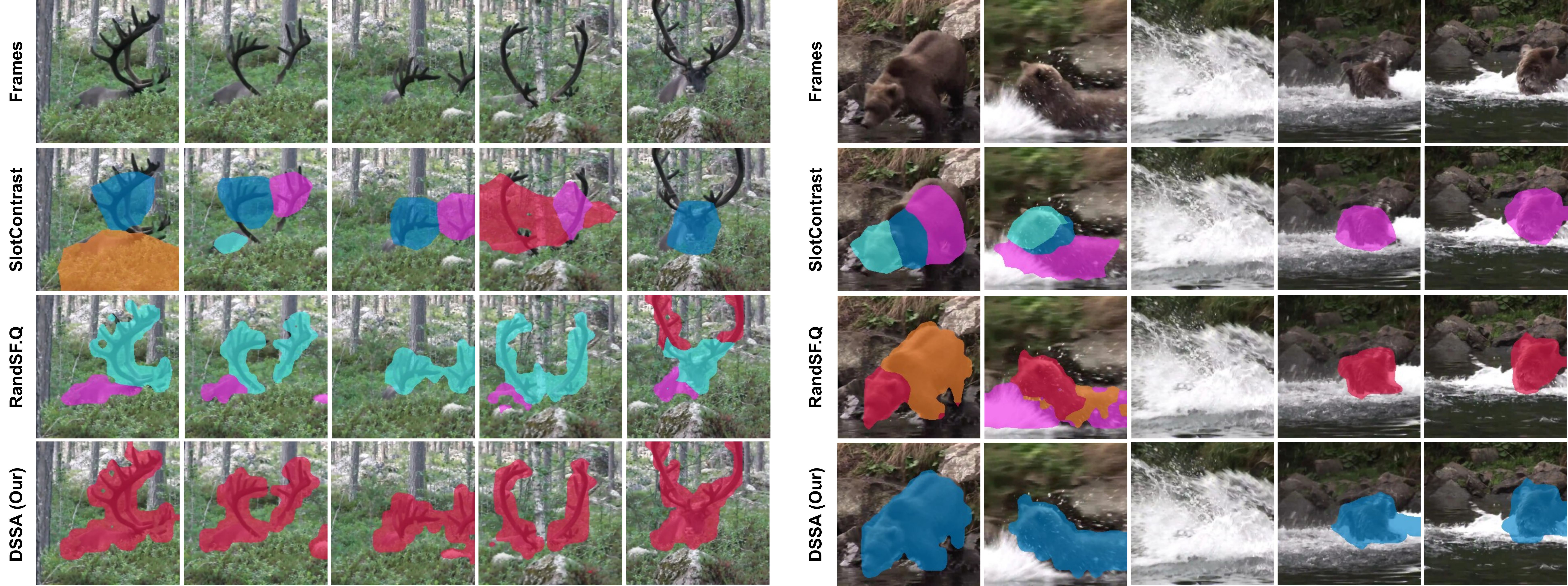}
  \vspace{-4mm}
  \caption{
    \textbf{Qualitative comparison on YTVIS.} 
    Colors denote slot identity; consistent colors across frames indicate 
    stable slot-to-object correspondence. Two challenging sequences are 
    shown: a deer moving through a forest under partial occlusion 
    (left), and a bear becoming fully occluded behind a waterfall before 
    re-emerging (right). DSSA maintains consistent slot identity 
    throughout both sequences, whereas SlotContrast~\cite{manasyan2025temporally} and RandSF.Q~\cite{zhao2025randsfq} exhibit 
    slot instability and fragmented object assignments.}
  \label{fig:qualitative_compare}
\vspace{-2mm}
\end{figure*}

\noindent
\textbf{Effect of $\alpha$ in CMA.}
Table~\ref{tab:alpha_ablation} reports ARI, ARIfg, mBO, and mIoU
across $\alpha \in \{0, 0.3, 0.5, 0.7, 1.0\}$ on MOVi-D.
At $\alpha=0$, CMA reduces to standard Slot Attention, where the
renormalization artifact causes slots to absorb spurious evidence from
unmatched tokens, reflected in lower ARI (37.7) despite competitive
ARIfg and mBO scores.
At $\alpha=1.0$, the modulation becomes too aggressive: aggregation
weights are scaled directly by the raw competitive weights, amplifying
inter-slot differences and destabilizing slot updates, which leads to
a notable drop across all metrics.
$\alpha = 0.5$ strikes the best balance, achieving the highest ARI (48.7) 
and ARIfg (82.9) while retaining strong mBO (30.1) and mIoU (28.5). 
This is consistent with the balanced interpolation interpretation in 
Section~\ref{sec:slot_attention}: values below 0.5 under-suppress inactive 
slots, while values above 0.5 over-penalize moderately attending ones.

\begin{table}[t]
  \centering
  \setlength{\tabcolsep}{10pt}
  \caption{
    \textbf{Effect of $\alpha$ in CMA} on MOVi-D. 
  }
  \label{tab:alpha_ablation}
  \vspace{-3mm}
  \resizebox{0.8\linewidth}{!}{%
  \begin{tabular}{c|cccc}
    \toprule
    $\alpha$ & ARI$\uparrow$ & ARIfg$\uparrow$ & mBO$\uparrow$ & mIoU$\uparrow$ \\
    \midrule
    $0.0$ & $37.7$ & $76.2$ & $25.8$ & $24.3$ \\
    $0.3$ & $35.8$ & $75.7$ & $25.0$ & $23.6$ \\
    $0.5$ & $\mathbf{48.7}$ & $\mathbf{82.9}$ & $\mathbf{30.1}$ & $\mathbf{28.5}$ \\
    $0.7$ & $47.6$ & $82.8$ & $29.7$ & $27.8$ \\
    $1.0$ & $44.5$ & $76.4$ & $27.8$ & $25.9$ \\
    \bottomrule
  \end{tabular}}
\vspace{-4mm} 
\end{table}

\subsection{Qualitative Results}
\label{sec:qualitative}
Figure~\ref{fig:qualitative_compare} compares slot assignment masks 
across five consecutive frames on YouTube-VIS for two challenging 
sequences. Colors denote slot identity; consistent colors across 
frames indicate stable slot-to-object correspondence. SlotContrast exhibits substantial slot instability across both 
sequences: the assigned colors change noticeably over time and the 
masks split the object into inconsistent parts, indicating weak 
slot-to-object correspondence under motion and partial occlusion. 
RandSF.Q improves temporal coherence, with several object parts 
tracked more consistently across frames, but still produces fragmented 
assignments and occasional ownership shifts--between the antlers, 
head, and body in the deer sequence (left), and between the head, 
body, and surrounding water regions in the bear sequence (right). 
In contrast, DSSA maintains a largely consistent slot identity 
throughout both sequences, assigning the same slot to the deer and 
bear across frames while preserving a more coherent object mask. 
These qualitative results suggest that DSSA better stabilizes slot 
ownership under appearance change and occlusion, which is consistent 
with its quantitative results in Table~\ref{tab:main}.

\subsection{Analysis: Two-Level Slot Representation}
\label{sec:analysis}

\begin{figure}[t]
  \centering
  \includegraphics[width=0.9\linewidth]{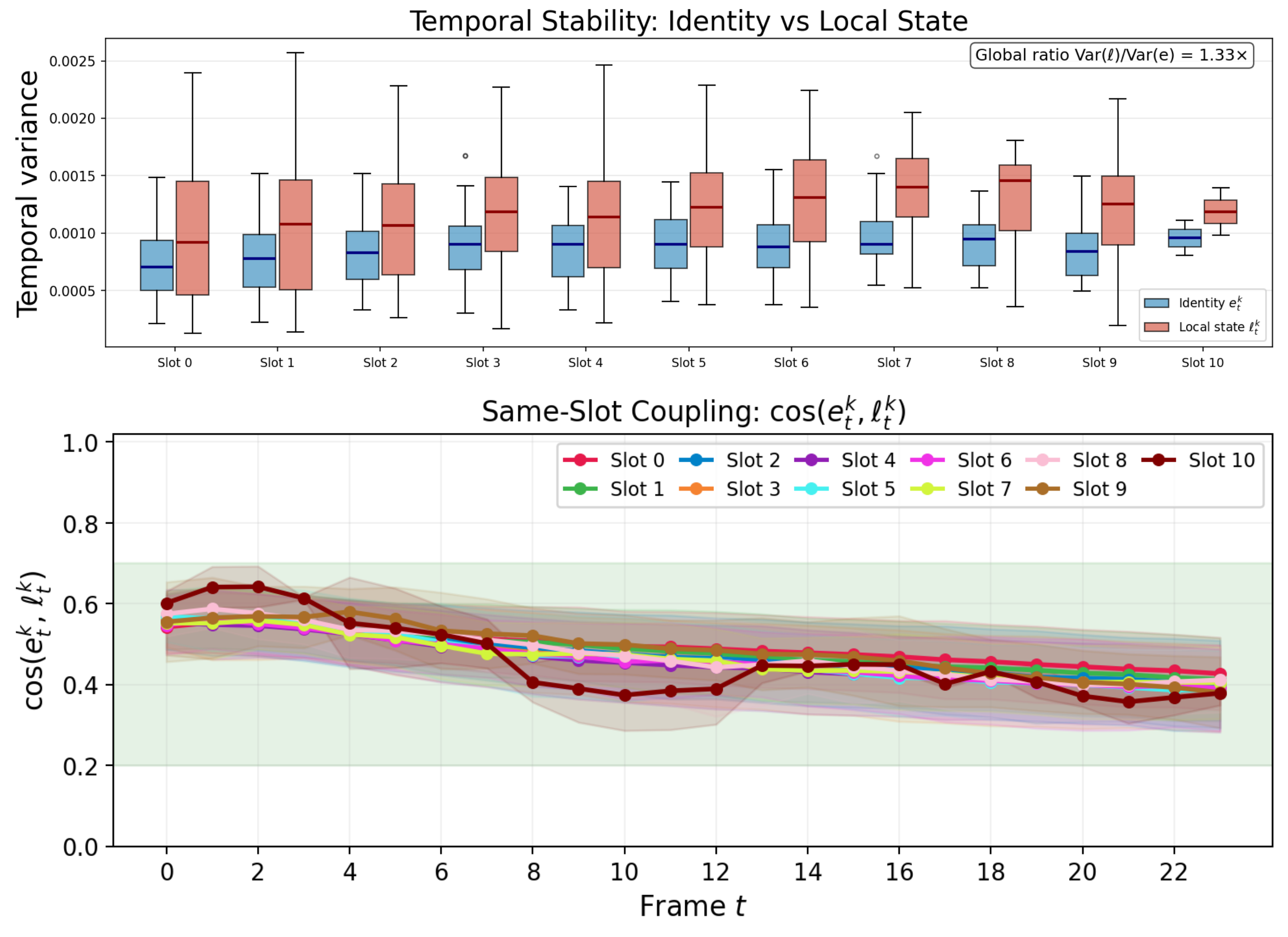}
  \vspace{-3mm}
  \caption{
    \textbf{Two-level slot representation analysis on MOVi-C.}
    \textit{Top}: temporal variance of $e_t^k$ (\textcolor{NavyBlue!60}{\rule{5pt}{1em}}) and $\ell_t^k$ (\textcolor{red!50}{\rule{5pt}{1em}}) 
    per active slot; local state $\ell_t^k$ exhibits higher variance for every slot 
    ($1.33\times$). \textit{Bottom}: cosine similarity $\cos(e_t^k,\ell_t^k)$ 
    over frames $t{=}0,\ldots,23$; per-slot mean of $0.473$ confirms 
    complementary information content.
  }
  \label{fig:two_level}
\vspace{-4mm}
\end{figure}

We verify empirically that the dual-state decomposition produces the
intended functional separation between the identity state $e_t^k$ and
local state $\ell_t^k$, measured over 250 evaluation sequences on
MOVi-C.

\smallskip
\noindent\textbf{Identity state is more temporally stable than local
state.} The top panel of Figure~\ref{fig:two_level} shows the temporal
variance of each representation, computed per-dimension and averaged
over $d$, for every active slot. The local state vectors $\ell_t^k$
exhibit consistently higher variance than the identity state vectors
$e_t^k$ across all 11 active slots, with a global ratio
$\overline{\mathrm{Var}_t(\ell^k)}\,/\,\overline{\mathrm{Var}_t(e^k)}
= 1.33$. The inter-quartile ranges do not overlap in any slot,
confirming the effect is systematic rather than driven by outliers.
 This result matches the intended roles of the two representations: $\ell_t^k$ is expected to respond to frame-specific changes such as appearance, pose, deformation, and occlusion, whereas $e_t^k$ should preserve the slowly varying information required to maintain object identity over time.

\smallskip
\noindent\textbf{The two representations remain distinct without
collapsing.} The bottom panel shows the cosine similarity
$\cos(e_t^k, \ell_t^k)$ over frames for each active slot. A 
per-slot mean of $0.473$ confirms that the two vectors do not overlap and carry complementary  information. The temporal flatness of these curves--declining by at most $0.2$ over the full sequence with narrow variance bands, indicating that the separation is maintained consistently
throughout the video rather than diminishing over time. Together, the two panels confirm a clear factorization: identity state and local state share sufficient structure to describe the same object while remaining complementary across~time.
\vspace{-2mm}
\section{Conclusion}
We presented Dual-State Slot Attention (DSSA), a self-supervised framework for video OCL that addresses two structural limitations of slot-based video OCL: (i) the representational conflict between reconstruction and temporal consistency, resolved by separating each slot into a local state for frame-specific appearance
and an identity state for temporally persistent object information, and (ii) the renormalization artifact of standard Slot Attention, resolved by competition-modulated aggregation (CMA). Experiments on MOVi-C, MOVi-D, and YouTube-VIS show consistent improvements over prior methods, with
gains up to $+9.7$ ARIfg on YouTube-VIS, while downstream evaluations confirm that the two states specialize in complementary ways for recognition and dynamics prediction. These findings highlight the
importance of explicitly disentangling appearance and identity for stable video OCL.

\noindent\textbf{Limitations and future work.} DSSA inherits several limitations from 
the slot-based OCL paradigm. The number of slots $K$ must be set in advance, and DSSA relies on a
frozen DINOv2 encoder, which may lack sufficient discriminability for visually similar or heavily occluded objects in unconstrained settings. Although DSSA improves temporal slot stability, challenging cases such as severe long-term occlusion and highly complex real-world dynamics
remain open problems. In future work, we aim to extend DSSA beyond the fixed $K$ and short-term temporal modeling by exploring adaptive slot allocation and longer-range memory, while evaluating on more diverse real-world benchmarks. The scope of downstream evaluation can be further broadened to tasks such as reasoning and embodied perception, along with extensions to 3D or multi-view settings.
\bibliographystyle{ACM-Reference-Format}
\bibliography{References}
\appendix


\section*{Supplementary material}

# Supplementary Material
# Dual-State Slot Attention: Decoupling Appearance and Identity for Video Object-Centric Learning

## 1 Experiment Details

*Main results.* For the main results reported in Table 1 of the main paper, we train DSSA using the Adam optimizer for 50k steps with a batch size of 8. The learning rate is set to $5 \times 10^{-5}$ for all datasets. We employ a learning rate schedule with a linear warmup over the first 2.5k steps, followed by cosine decay, and apply gradient clipping with a maximum norm of 0.05. During training, we sample short video clips of consecutive frames, with a segment length of 6 for MOVi-C and MOVi-D, and 5 for YTVIS. All experiments use a frozen DINOv2 ViT-S/14 encoder with input resolution $256 \times 256$, producing 256 image tokens of dimension 384. We apply random horizontal flipping as data augmentation. The number of slots is set to 11, 21, and 7 for MOVi-C, MOVi-D, and YTVIS, respectively. Slot Attention uses 3 iterations for the first frame and 1 iteration for subsequent frames. The slot dimensionality is 256 for MOVi-C/MOVi-D and 64 for YTVIS. The CMA coefficient $\alpha$ is fixed to 0.5. The transition module $\mathcal{T}$ is implemented as a Transformer decoder layer with 4 attention heads, and FFN dimensions of 1024, 1024, and 256 for MOVi-C, MOVi-D, and YTVIS, respectively. The slot decoder reconstructs image features from slot representations while predicting per-slot masks, and is implemented as a 4-layer Transformer decoder with 4 attention heads and an FFN dimension of 1,536. Complete hyperparameter configurations are provided in Table S-I.

**Table S-I: Full hyperparameters of DSSA for all three benchmarks. All models use a frozen DINOv2 ViT-S/14 encoder and $256 \times 256$ input resolution.**

| Hyperparameter | Benchmarks | | |
|---|---|---|---|
| | MOVi-C | MOVi-D | YTVIS |
| *Optimization* | | | |
| Optimizer | Adam | | |
| Training steps | 50,000 | | |
| Batch size | 8 | | |
| Training segment length | 6 | 6 | 5 |
| Initial learning rate | 5e-5 | | |
| LR warmup steps | 2,500 | | |
| Gradient norm clipping | 0.05 | | |
| *Image specification* | | | |
| ViT architecture | DINOv2 ViT-S/14 | | |
| Feature dim | 384 | | |
| Image tokens $N$ | 256 | | |
| Augmentations | Random horizontal flip | | |
| *Slot Attention / CMA* | | | |
| Number of slots $K$ | 11 | 21 | 7 |
| SA iterations (1st / subsequent frames) | 3 / 1 | | |
| Slot dimension $d$ | 256 | 256 | 64 |
| CMA modulation $\alpha$ (Eq.5) | 0.5 | | |
| *Transitioner $\mathcal{T}$* | | | |
| Type | Transformer decoder layer | | |
| Attention heads / FFN dim | 4 / 1024 | 4 / 1024 | 4 / 256 |
| *Decoder* | | | |
| Type | Transformer decoder | | |
| Layers / Heads / FFN dim | 4 / 4 / 1,536 | | |

*Downstream Tasks.* We further evaluate the learned representations via two downstream tasks on YTVIS, assessing the local state $\ell_t^k$ and identity state $e_t^k$ separately to verify that each has specialized as intended. These tasks probe complementary aspects of the learned representations: object recognition tests whether slots encode discriminative per-frame appearance information, while object dynamics prediction tests whether slots capture temporally stable identity information suitable for forecasting future object states.

**Object recognition.** Following RandSF.Q [9], we train a two-layer MLP on top of frozen DSSA representations, keeping the entire DSSA backbone frozen throughout this stage. We train for 5,000 steps with a batch size of 8. The MLP has an input dimension of $D_{\text{slot}}$ and an output dimension of 45 (41 classes and 4 bounding box coordinates), with pre-layer normalization and a dropout rate of 0.1. We use a learning rate of $1 \times 10^{-3}$.

**Object dynamics prediction.** Following SlotContrast [4], we train SlotFormer [7] on top of frozen DSSA representations (Section 4.3, main paper), keeping the entire DSSA backbone frozen throughout this stage. We train for 100,000 steps with a batch size of 16. The temporal configuration uses $T = 10$ burn-in frames and $K = 5$ rollout frames. Each slot is encoded with a latent size of 128, and the Transformer backbone consists of 4 layers with a dropout rate of 0.1. We use a peak learning rate of $5 \times 10^{-5}$.

*Hardware.* All experiments are run on NVIDIA A6000 GPUs.

## 2 Metrics Details

We evaluate all methods using four video-level metrics computed over full video sequences. Computing metrics at the video level – rather than per frame – is essential for capturing temporal consistency: a method that produces accurate per-frame masks but swaps slot identities across frames will be appropriately penalized.

***ARI and ARI$_{fg}$*** . The Adjusted Rand Index (ARI) measures clustering agreement between predicted and ground-truth masks across all pixels and frames of a video, penalizing both incorrect spatial assignments and temporal inconsistencies such as slot swaps and identity drift. ARI$_{\text{fg}}$ is a variant computed by excluding background pixels, focusing on foreground object segmentation quality.

***mBO.*** Mean Best Overlap measures, for each ground-truth mask, the highest IoU against all predicted masks, averaged across all objects and frames. It assesses mask sharpness and spatial coverage without requiring consistent slot assignment across frames.

***mIoU.*** Mean Intersection-over-Union requires consistent slot-to-object assignment rather than just the best local match, making it a stricter spatial accuracy measure than mBO.

***On the limitations of standalone metrics and the importance of balanced evaluation.*** While each metric captures a meaningful aspect of segmentation quality, relying on any single metric in isolation can be misleading. Several prior works have raised concerns specifically about $\text{ARI}_{\text{fg}}$ as a standalone metric [1–3, 5, 6, 8]: its exclusion of background pixels limits its ability to fully capture segmentation quality, and it has been shown to favor over- or under-segmentation in certain settings. More broadly, we observe that each metric has its own blind spot that can be exploited by degenerate predictions.

To illustrate this, Figure S-I presents three simulated failure cases (Predictions B–D) alongside our model output (Prediction A), with quantitative results reported in Table S-II.

***Prediction B – All slots collapsed.*** The entire image is represented by a single slot, producing no meaningful scene decomposition. Despite this complete failure, $\mathbf{ARI_{fg} = 1.0}$ – an artifact of the metric definition: when all pixels belong to one cluster, all foreground pixels trivially share the same assignment. ARI is similarly inflated by the dominant background area. Only mBO and mIoU correctly reflect the failure, both at 0.147.

***Prediction C – Foreground over-segmentation.*** The foreground object is fragmented into an excessive number of segments while the background is handled reasonably. **ARI = 0.939**, the highest among all cases, is artificially boosted by the well-segmented background and masks the foreground failure. $\text{ARI}_{\text{fg}}$ drops to 0.455, partially capturing the fragmentation, but mBO and mIoU more directly penalize the over-segmentation at 0.381.

***Prediction D – Background over-segmentation.*** The foreground is well-captured while the background is split into many spurious segments. $\mathbf{ARI_{fg} = 0.958}$ appears near-perfect, yet **ARI = 0.310** drops sharply due to the fragmented background. Notably, mBO remains relatively high at 0.770, since per-segment overlap can still be large even when the number of background segments is excessive – illustrating that mBO alone is also insufficient in this regime.

These cases demonstrate that each metric can be individually deceived by a specific failure mode. In contrast, our model output (Pred. A) achieves a consistently strong and balanced profile across all four metrics, without being artificially inflated on any single one. We argue that this balanced performance is a more reliable indicator of genuine segmentation quality than peak scores on individual metrics, which — as shown above — can be achieved trivially by degenerate predictions. Consistent with this observation and with prior concerns about $\text{ARI}_{\text{fg}}$ [1, 2], we report all four metrics throughout this work and, when methods differ across them, place greater weight on mBO and mIoU as more reliable indicators, treating ARI and $\text{ARI}_{\text{fg}}$ as complementary signals rather than primary criteria.

**Table S-II: Quantitative results for three simulated failure cases and our model output (Prediction A). †Simulated cases constructed to isolate specific failure modes.**

| | **ARI↑** | **$\text{ARI}_{\text{fg}}$↑** | **mBO↑** | **mIoU↑** |
|---|---|---|---|---|
| Prediction B: Slot collapse† | 0.857 | 1.000 | 0.147 | 0.147 |
| Prediction C: FG over-seg† | 0.939 | 0.455 | 0.381 | 0.381 |
| Prediction D: BG over-seg† | 0.311 | 0.958 | 0.770 | 0.770 |
| Prediction A (ours) | 0.916 | 0.718 | 0.591 | 0.591 |

# 3 Additional Ablation Studies

## 3.1 Loss Component Ablation

Table S-III isolates the contribution of each of the three training objectives that act on distinct components of the dual-state architecture: the reconstruction loss $\mathcal{L}_{\text{recon}}$ (Eq. 8) on the local states $\{\ell_t^k\}$, the auxiliary reconstruction loss $\mathcal{L}_{\text{aux}}$ (Eq. 9) which grounds the identity states in the object's visual properties, and the temporal identity consistency loss $\mathcal{L}_{\text{id}}$ (Eq. 10) on the identity states $\{e_t^k\}$. All rows use the full dual-state architecture; only the supervision signals differ.

The results reveal two complementary failure modes. First, Exp. #2 removes $\mathcal{L}_{\text{aux}}$ while retaining $\mathcal{L}_{\text{id}}$, yielding the highest ARI (74.4) but causing a sharp degradation in spatial quality ($\text{ARI}_{\text{fg}}$ 43.9, mBO 19.2, mIoU 16.7). Without a reconstruction anchor, $\mathcal{L}_{\text{id}}$ can push the identity states toward temporal consistency without being constrained to remain visually meaningful, resulting in poorly localized slot masks. Second, Exp. #3 removes $\mathcal{L}_{\text{id}}$ while retaining $\mathcal{L}_{\text{aux}}$, which preserves spatial quality ($\text{ARI}_{\text{fg}}$ 65.9, mBO 26.3) but leads to a notable drop in ARI (56.8), as slot swapping is no longer explicitly suppressed. The full model (Exp. #4) resolves both failure modes simultaneously: $\mathcal{L}_{\text{id}}$ provides temporal direction while $\mathcal{L}_{\text{aux}}$ ensures the identity state remains grounded in object appearance, together achieving the best balance across all four metrics.

**Table S-III: Loss component ablation on MOVi-C.**

| Exp. | $\mathcal{L}_{\text{recon}}$ | $\mathcal{L}_{\text{id}}$ | $\mathcal{L}_{\text{aux}}$ | ARI↑ | $\text{ARI}_{\text{fg}}$↑ | mBO↑ | mIoU↑ |
|---|---|---|---|---|---|---|---|
| #1 | ✓ | | | 49.2 | 62.3 | 22.1 | 20.3 |
| #2 | ✓ | ✓ | | 74.4 | 43.9 | 19.2 | 16.7 |
| #3 | ✓ | | ✓ | 56.8 | 65.9 | 26.3 | 24.1 |
| #4 | ✓ | ✓ | ✓ | **68.6** | **67.6** | **28.9** | **26.5** |

## 3.2 Stop-Gradient Ablation

The stop-gradient $\text{sg}(\ell_t^k)$ in Eq. (7) is the key mechanism that structurally decouples the gradient paths of the two states. Without it, $\mathcal{L}_{\text{id}}$ and $\mathcal{L}_{\text{aux}}$ back-propagate into the local state, allowing identity-related objectives to interfere with per-frame appearance learning. Table S-IV shows that removing the stop-gradient leads to a consistent drop across all four metrics (ARI: 68.6 → 56.9, $\text{ARI}_{\text{fg}}$ : 67.6 → 60.2, mBO: 28.9 → 26.5, mIoU: 26.5 → 24.9), confirming that gradient decoupling is necessary for both temporal consistency and spatial segmentation quality.

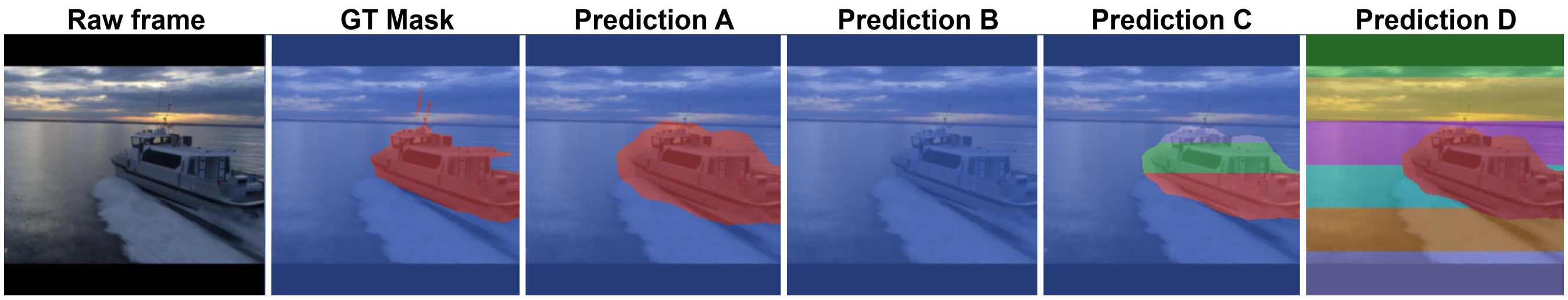


**Figure S-I: Visualization of three simulated failure cases together with our model output. From left to right: (a) input image, (b) ground-truth mask, (c) Prediction A (our model output), (d) Prediction B: all slot collapse, (e) Prediction C: foreground over-segmentation, and (f) Prediction D: background over-segmentation. The simulated failure cases each achieve a deceptively high score on at least one metric while failing on others, motivating joint evaluation across multiple metrics.**

**Table S-IV: Effect of stop-gradient on $\ell_t^k$ (Eq. 7, main paper) on MOVi-C.**

| **Variant** | ARI↑ | $\text{ARI}_{\text{fg}}$↑ | mBO↑ | mIoU↑ |
|---|---|---|---|---|
| DSSA w/o stop-gradient | 56.9 | 60.2 | 26.5 | 24.9 |
| DSSA w/ stop-gradient (ours) | **68.6** | **67.6** | **28.9** | **26.5** |

## 4 Occluded-Object Experiment

A key motivation of DSSA is robustness to occlusion (Section 1, main paper). To directly evaluate this, we construct a targeted MOVi-C subset containing only sequences in which at least one object (i) appears with visibility $\geq n_{\min}$ pixels, (ii) becomes *fully* occluded (visibility = 0), and (iii) re-emerges with visibility $\geq n_{\min}$ pixels. We set $n_{\min} = 400$ to exclude very small objects, yielding 53 sequences.

Table S-V reports results on this subset. DSSA substantially outperforms baselines, particularly in video-level ARI, which directly penalises failed re-identifications after occlusion. The identity state $e_t^k$, maintained through the stop-gradient GRU, serves as a latent memory that anchors the slot to its object during the occluded interval.

**Table S-V: Performance on the MOVi-C fully-occluded subset.**

| **Method** | ARI↑ | $\text{ARI}_{\text{fg}}$↑ | mBO↑ | mIoU↑ |
|---|---|---|---|---|
| SlotContrast [4] | 47.4 | 64.4 | 21.6 | 19.9 |
| RandSF.Q [9] | 69.1 | 65.0 | 28.6 | 26.4 |
| **DSSA (ours)** | **71.3** | **70.1** | **30.1** | **28.2** |

## 5 Additional Qualitative Results

- **Figure S-II:** Qualitative comparison between DSSA and RandSF.Q on YTVIS.
- **Figure S-III:** Qualitative comparison between DSSA and RandSF.Q on MOVi-C and MOVi-D.

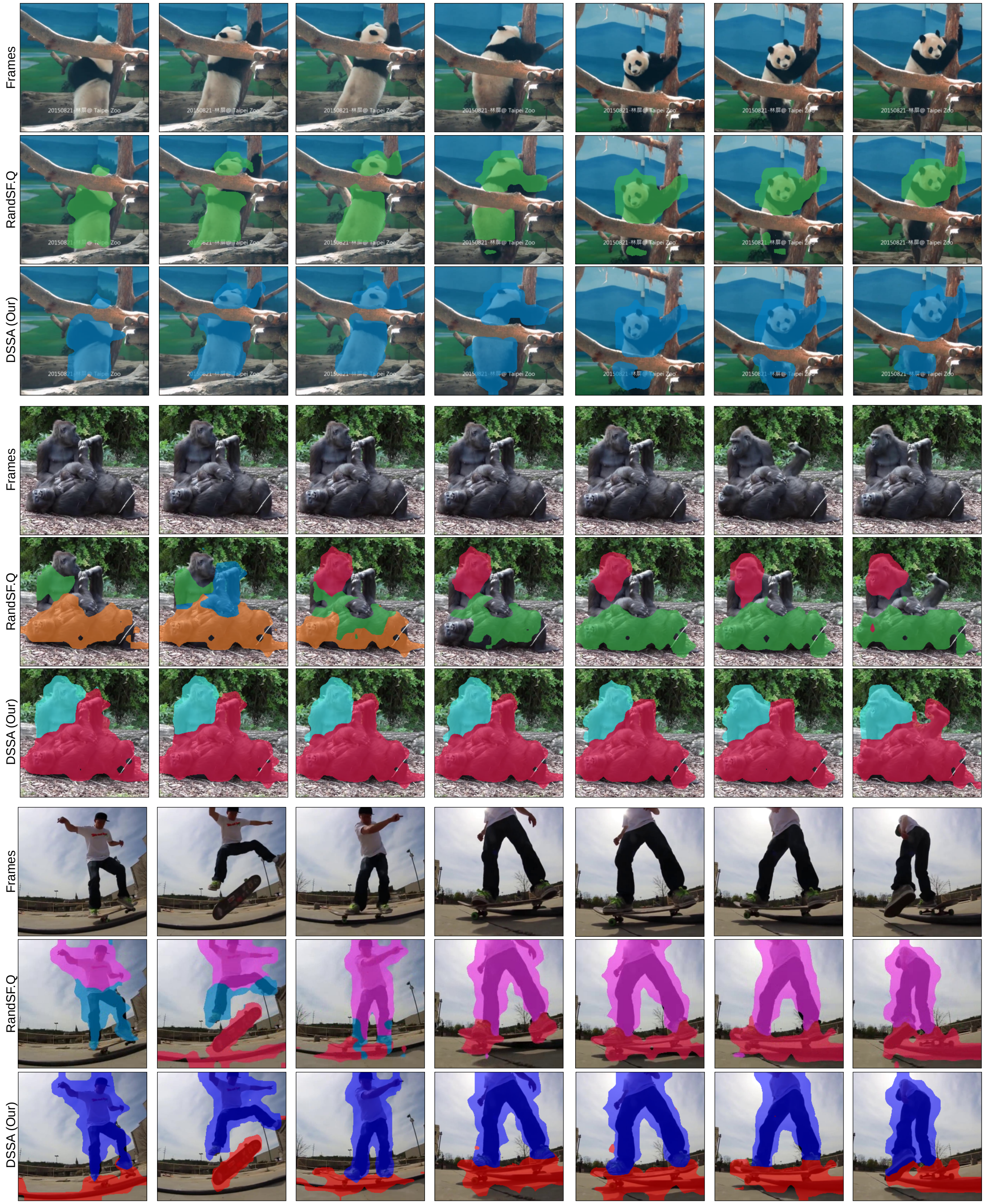


**Figure S-II: Additional qualitative comparison on YTVIS.**

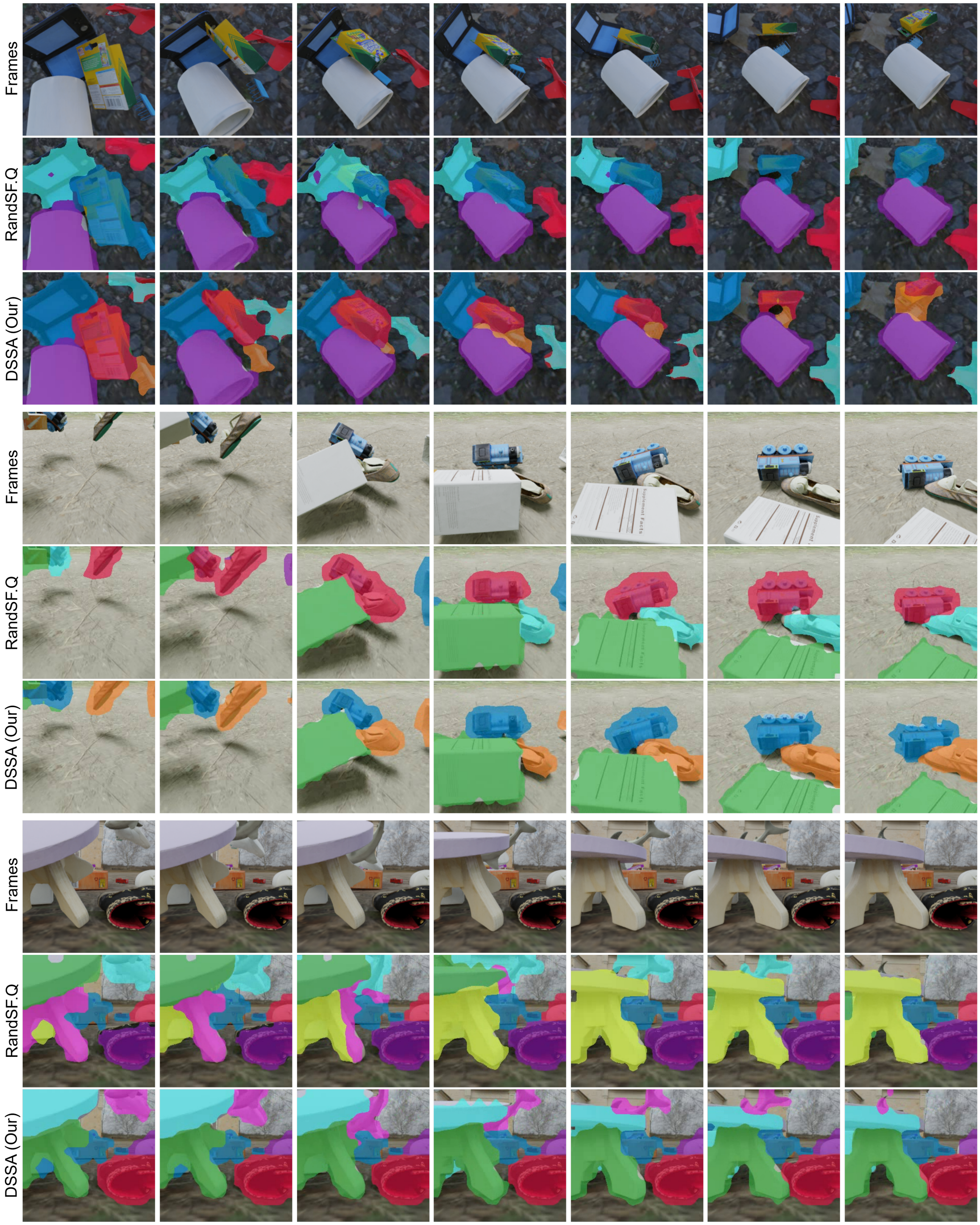


**Figure S-III: Additional qualitative comparison on MOVi-C and MOVi-D.**



\end{document}